\pdfoutput=1   % arXiv: must stay within the first 5 lines
\documentclass[11pt]{article}
\usepackage[final]{lucas-arxiv}

\makeatletter
\def\input@path{{./}{../}}
\makeatother
\graphicspath{{./}{../}{figures/}{../figures/}}
\usepackage{capt-of}

\newcommand{\draftnote}[1]{}

\newcommand{\promptfont}{\fontfamily{zi4}\selectfont\footnotesize\frenchspacing\raggedright}
\newtcolorbox{promptcard}[2][]{enhanced, colback=gray!3, colframe=gray!55,
	colbacktitle=gray!15, coltitle=black, fonttitle=\small\bfseries,
	title={#2}, boxrule=0.5pt, arc=2pt, left=5pt, right=5pt, top=4pt, bottom=4pt,
	toptitle=1.5pt, bottomtitle=1.5pt, before skip=6pt, after skip=8pt,
	fontupper=\promptfont, fontlower=\promptfont,
	segmentation style={gray!55, dashed}, breakable, #1}
\newcommand{\promptslot}[1]{\textcolor{blue!55!black}{#1}}
\newcommand{\promptrole}[1]{{\normalfont\scriptsize\scshape\textcolor{gray!85!black}{#1}}\par\smallskip}
\newcommand{\promptskip}{\par{\textcolor{gray!70}{[\ldots]}}\par}
\newtcolorbox{promptchanged}{enhanced, sharp corners, boxrule=0pt, frame hidden,
	colback=blue!7, borderline west={1.5pt}{0pt}{blue!55!black},
	left=4pt, right=2pt, top=1pt, bottom=1pt, before skip=2pt, after skip=2pt,
	fontupper=\promptfont}

\logos{\includegraphics[height=\logoheight]{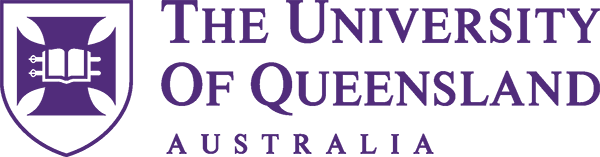},
       \includegraphics[height=\logoheight]{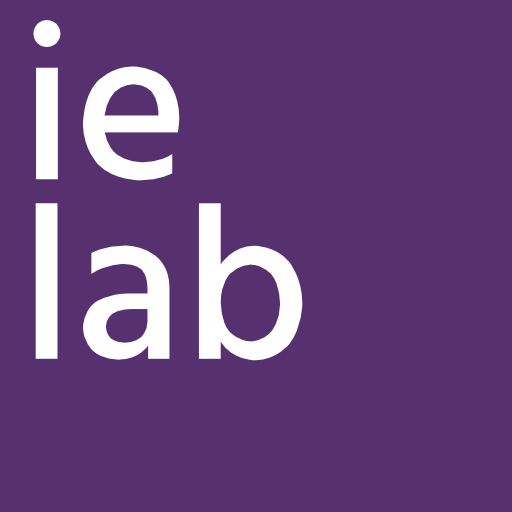}}
\title{Do Conversational Agents Need Explicit Memory?\\A Controlled Comparison with Raw-History Retrieval}
\underreview   % or \accepted[<DOI>]{TACL} once accepted
\author{Yongjie Zhou\textsuperscript{1} \and Shuai Wang\textsuperscript{1}
  \and Bevan Koopman\textsuperscript{1,2} \and Guido Zuccon\textsuperscript{1,3}}
\affiliation{\textsuperscript{1}The University of Queensland, Brisbane, Australia
  \quad \textsuperscript{2}CSIRO \quad \textsuperscript{3}Google}
\email{\{yongjie.zhou, shuai.wang2, g.zuccon\}@uq.edu.au, bevan.koopman@csiro.au}
\codelink{https://github.com/ielab/memory-systems-vs-rag}
\date{}
\hypersetup{pdftitle={Do Conversational Agents Need Explicit Memory? A Controlled Comparison with Raw-History Retrieval},
  pdfauthor={Yongjie Zhou, Shuai Wang, Bevan Koopman, Guido Zuccon}}

\begin{document}
\maketitle

\begin{abstract}
Explicit memory systems transform conversation history into persistent representations intended to improve future access, often at substantial construction cost. Yet it remains unclear whether this added complexity yields better answer accuracy than retrieving directly from raw history. This study compares six state-of-the-art explicit memory systems with raw-history RAG on LongMemEval-S and BEAM-100K under original and controlled retrieval settings. We find only localized gains for particular question types and tight run-time context budgets, with no consistent advantage over strong raw-history RAG baselines. Oracle retrieval shows that some systems fail to preserve answer-supporting information during explicit memory construction, while others preserve it but fail to retrieve it effectively, despite construction costs reaching hours per conversation.

Our findings suggest that while many existing memory systems for AI agents emphasize increasingly elaborate explicit memory construction, they are often evaluated without strong controlled comparisons to raw-history RAG and may underemphasize retrieval. Retrieval remains a major bottleneck when useful information is preserved, suggesting that improving access to existing information may be at least as important as constructing richer explicit memory.
\end{abstract}

\section{Introduction}

AI agents that interact with users over long periods need to recover relevant information from a growing conversation history. Recent memory systems address this challenge by transforming past user--assistant interactions into persistent representations, such as summaries, extracted facts, and linked notes, that can later be retrieved when needed \citep{zhong2023memorybank,xu2025amem,chhikara2025mem0,fang2025lightmem,liu2026simplemem,yue2026hypermem}. We refer to these approaches as ``\emph{explicit memory systems}'', and to the resulting representations as ``\emph{explicit memory}''. Constructing and maintaining such memory, however, can require substantial additional computation, including repeated LLM calls for extraction, summarization, consolidation, linking, or reflection. A natural alternative is to avoid this construction stage altogether and retrieve relevant evidence directly from the original conversation history. We refer to this alternative as \emph{raw-history retrieval-augmented generation (RAG)}. This raises a fundamental question: \emph{does explicit memory construction provide enough benefit to justify its additional cost over raw-history RAG?}

Evaluating this trade-off is not straightforward. The effectiveness of an explicit memory system depends not only on what is constructed, but also on how the memory is retrieved, how much retrieved memory is supplied to the answer model, and how much computation is required throughout the pipeline. Consequently, end-to-end comparisons between systems cannot determine whether observed gains come from explicit memory construction itself, from differences in retrieval and context budget, or from additional computation.

In this study, we examine this trade-off through a controlled evaluation of six state-of-the-art explicit memory systems against raw-history RAG. We ask: (1) whether explicit memory improves answer accuracy over raw-history RAG; (2) whether its limitations arise from explicit memory construction or retrieval; and (3) whether any resulting gains justify the additional computation required to construct and use explicit memory. Beyond evaluating each system under its original configuration, we hold its constructed explicit memory fixed and systematically vary the retriever and context budget under a unified answering setup. We further use oracle retrieval to separate limitations in explicit memory construction from those in retrieval, and measure both construction and query-time cost.

% Vector TikZ figure; source in figures/src/fig1_framework.tex.
\begin{figure*}
\centering
\includegraphics[width=\textwidth]{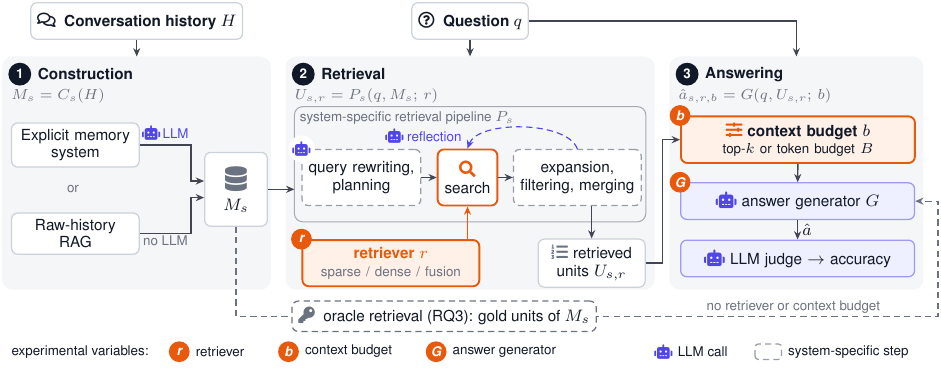}
\caption{Experimental framework. Explicit memory systems and raw-history RAG share a three-stage evaluation pipeline of construction, retrieval, and answering (Section~\ref{sec:evaluation_pipeline}).}

\label{fig:framework}
\end{figure*}

Our results show that the evaluated explicit memory systems provide no consistent overall advantage over raw-history RAG. This pattern persists when retrieval and context budgets are controlled, although explicit memory provides localized gains for some question types and under tight context budgets. Oracle retrieval reveals two distinct bottlenecks: some systems preserve useful information but fail to retrieve it effectively, whereas others appear to lose useful information during explicit memory construction. Moreover, the additional construction and query-time computation required by explicit memory does not consistently translate into higher answer accuracy.

%Taken together, 
Our findings suggest that improving conversational memory may require shifting attention from increasingly elaborate memory construction toward retrieval. Across the evaluated systems, explicit memory construction yields limited and inconsistent gains, while retrieval remains a major bottleneck even when useful information is preserved. This suggests that better access to existing information may be at least as important as constructing richer memory representations.
\section{Background and Motivation}
\label{sec:motivation}

Long-term conversational memory systems differ widely in how they construct and retrieve memory. Early agent architectures such as Generative Agents and MemGPT explored persistent memory and selective retrieval \citep{park2023generative,packer2023memgpt}, while more recent work increasingly constructs explicit memory from conversation history. 
MemoryBank combines conversation content with summaries; Mem0 and LightMem extract compact memory units; A-Mem constructs linked notes; SimpleMem combines structured memory with query-aware retrieval; and HyperMem organizes memory hierarchically using a hypergraph \citep{zhong2023memorybank,chhikara2025mem0,xu2025amem, fang2025lightmem,liu2026simplemem,yue2026hypermem}. We focus on these six explicit memory systems because they construct persistent explicit memory and use a retrieval stage that can be studied separately from construction. Their different construction and retrieval procedures make end-to-end results difficult to interpret.

A system that performs better may construct more useful explicit memory, retrieve from it more effectively, provide more relevant information to the answer model, or simply use more computation. Conversely, a system that performs poorly may be limited by explicit memory construction, retrieval, or both. These ambiguities reveal four gaps in current evaluation practice.

\textbf{Gap 1: End-to-end comparisons do not isolate the value of explicit memory construction.} 
Recent explicit memory systems are typically evaluated as complete pipelines, though some recent work has begun to separate memory construction from retrieval \citep{zhou2026reproducing,hu2026does}.
A-Mem, SimpleMem, and HyperMem, for example, jointly change both the constructed explicit memory and retrieval \citep{xu2025amem,liu2026simplemem,yue2026hypermem}.
Mem0 and LightMem include simpler RAG-style baselines \citep{chhikara2025mem0,fang2025lightmem}, but the compared systems still differ in retrieval procedure, context budget, or answering setup. An end-to-end gain therefore does not establish that explicit memory construction itself is responsible for the improvement.

%\vspace{0.5em}
%\problembox{Gap 2: Retrieval is rarely controlled across explicit memory systems.}
%\vspace{0.5em}

\textbf{Gap 2: Retrieval is rarely controlled across explicit memory systems.}
The evaluated systems use very different retrieval procedures, ranging from dense retrieval to link expansion, query planning, hybrid search, and hierarchical retrieval. Because retrieval determines which preserved information reaches the answer model, it can materially change performance even when the constructed explicit memory is fixed. A controlled comparison must therefore evaluate explicit memory systems under matched retriever and context-budget settings.

\textbf{Gap 3: End-to-end accuracy cannot distinguish retrieval failure from information lost during explicit memory construction.} 
A poor answer can arise because relevant information was discarded during explicit memory construction, or because it remains in explicit memory but is not retrieved. These failures have different causes and require different solutions, but standard end-to-end evaluation cannot distinguish them. We use oracle retrieval to measure the answer accuracy supported by each system's explicit memory when retrieval is bypassed.

\textbf{Gap 4: Answer accuracy alone does not capture the cost of explicit memory.} 
Explicit memory systems may require substantial LLM computation for extraction, summarization, consolidation, linking, query planning, or reflection. Several recent systems explicitly optimize for efficiency \citep{fang2025lightmem,liu2026simplemem}, but the relevant comparison is also against raw-history RAG, which requires no explicit memory construction. We therefore evaluate construction and query-time cost together with answer quality.
\section{Experimental Framework}
\label{sec:experimental_framework}

To address the evaluation gaps in Section~\ref{sec:motivation}, we place all systems within a common three-stage pipeline: construction, retrieval, and answering. The central principle is to separate explicit memory construction from downstream retrieval and answering. Constructed explicit memory remains fixed, while the retriever, context budget, and answering setup are varied according to the research question. Figure~\ref{fig:framework} summarizes the framework. Sections~\ref{sec:rq1}--\ref{sec:rq4} address four research questions: answer accuracy under original configurations (RQ1), sensitivity to retrievers, context budgets, and answer generators (RQ2), oracle answer accuracy (RQ3), and computational cost (RQ4).

\subsection{Evaluation Pipeline}
\label{sec:evaluation_pipeline}

Let $H$ denote a conversation history, $q$ a question about that history, and $s$ an explicit memory system or raw-history RAG baseline. A \emph{retrieval unit} is an item that can be returned by a retrieval pipeline, such as a memory unit, conversation turn, or user--assistant turn pair.

\paragraph{Construction.}
System $s$ constructs a collection of retrieval units
$M_s = C_s(H)$,
where $C_s$ denotes its system-specific construction procedure. For explicit memory systems, $C_s$ may summarize, extract, update, or organize conversation content into explicit memory. For raw-history RAG, $C_s$ directly indexes the original conversation history without LLM-based memory construction. Each collection $M_s$ is constructed once and remains fixed throughout the controlled experiments.

\paragraph{Retrieval.}
Given question $q$, system $s$ retrieves units
$U_{s,r} = P_s(q, M_s; r)$
from $M_s$, where $P_s$ denotes the full system-specific retrieval pipeline and $r$ the retriever used within that pipeline. Depending on the system, $P_s$ may additionally include query rewriting, planning, multiple searches, structural expansion, filtering, merging, deduplication, or reflection. In the controlled experiments, we vary $r$ while preserving the surrounding system-specific retrieval logic.

\paragraph{Answering.}
Retrieved units are included in the answer prompt subject to context budget $b$, and the answer is generated as
$\hat{a}_{s,r,b} = G\!\left(q, U_{s,r}; b\right)$,
where $G$ denotes the answer generator and $\hat{a}_{s,r,b}$ the generated answer. The notation indicates that $G$ receives the question together with retrieved units selected under context budget $b$, using the answer prompt specified for the experiment. We consider two forms of context budget: a maximum number of retrieved units $k \in \{3,5,10,20\}$ and a token budget $B \in \{200,500,1000,2000\}$.

\subsection{Systems and Baselines}
\label{sec:systems}

We evaluate the six explicit memory systems introduced in Section~\ref{sec:motivation}: MemoryBank \citep{zhong2023memorybank}, A-Mem \citep{xu2025amem}, Mem0 \citep{chhikara2025mem0}, LightMem \citep{fang2025lightmem}, SimpleMem \citep{liu2026simplemem}, and HyperMem \citep{yue2026hypermem}. Table~\ref{tab:app_adaptations} in Appendix~\ref{app:systems} summarizes their construction, retrieval, and answering components.

As controls without explicit memory construction, we evaluate raw-history RAG with three retrieval-unit definitions: \emph{all-turns}, containing all conversation turns; \emph{user-turns}, containing user turns only; and \emph{pair-turns}, containing user--assistant turn pairs. These baselines use the same conversation histories while varying speaker coverage and retrieval granularity.
We also include a \emph{full-context} baseline that bypasses retrieval and supplies raw conversation history directly to the answer generator. The history is capped at 100K tokens\footnote{The answer generator supports up to 138K context tokens; we use a lower cap to leave room for the question, prompt, and generated answer.}; if it exceeds this limit, we retain the most recent content that fits.

\subsection{Datasets and Evaluation}
\label{sec:datasets}

We evaluate on LongMemEval-S \citep{wu2024longmemeval} and BEAM-100K \citep{tavakoli2025beam}. LongMemEval-S contains 500 questions, each paired with its own conversation history, testing five long-term memory abilities: information extraction, multi-session reasoning, knowledge updates, temporal reasoning, and abstention. BEAM-100K contains 20 conversations, each with 20 questions (400 in total), spanning a broader set of conversational memory abilities, including knowledge updating, multi-session reasoning, instruction and preference following, summarization, temporal reasoning, and abstention.

\paragraph{Metrics.}
We report accuracy (\%) on both benchmarks using their respective scoring criteria. For LongMemEval-S, accuracy is the percentage of questions judged correct according to the benchmark's question-type-specific criteria. For BEAM-100K, accuracy is the mean per-question benchmark score, providing a graded measure of answer correctness; most question types use rubric-based scoring, while event ordering uses normalized Kendall's $\tau_b$ multiplied by event-matching F1. We use GPT-4o mini to evaluate answers from all systems according to these benchmark-specific criteria. For BEAM-100K, we correct the released judging setup by restoring the question to the judge prompt and clarifying that responses explicitly stating the absence of relevant information should be treated as responsive in abstention cases (Appendix~\ref{app:judging}). The corrected judging setup is applied consistently across all BEAM-100K experimental conditions.

\subsection{Models and Answering Setup}
\label{sec:generation}

Unless otherwise specified, we use Qwen3-30B-A3B-Instruct-2507 \citep{qwen3technicalreport} for LLM-based memory construction and answer generation, served with vLLM.

Section~\ref{sec:rq1} (RQ1) preserves each system's original retrieval and answering setup. In Sections~\ref{sec:rq2} and~\ref{sec:rq3} (RQ2--RQ3), all systems instead use a unified answering setup with the same answer prompt (Appendix~\ref{app:prompts}) and answer generator. RQ2 varies the retriever and context budget, and additionally varies the answer generator as a robustness check. RQ3 replaces practical retrieval with oracle retrieval.
\section{RQ1: Do Explicit Memory Systems Outperform Raw-History RAG?}
\label{sec:rq1}

% Revised from _paper/scripts/build_rq_tables.py output.
% Apply the corresponding formatting changes to the generator before regeneration.
\begin{table}
\centering
\footnotesize
\setlength{\tabcolsep}{2.5pt}
\begin{tabular}{@{}llcc@{}}
\toprule
Method & Retrieval & LME-S & BEAM \\
\midrule
Full context & --- & 22.2 & 24.4 \\
\midrule
\multicolumn{4}{@{}l}{\textit{Raw-history RAG (unified prompt)}} \\
\emph{all-turns}
 & dense (MiniLM) & 63.2 & \textbf{56.2} \\
 & dense (Qwen3-Emb-0.6B) & \textbf{78.0} & 56.0 \\
\emph{user-turns}
 & dense (MiniLM) & 65.6 & 54.8 \\
 & dense (Qwen3-Emb-0.6B) & 72.0 & 55.9 \\
\emph{pair-turns}
 & dense (MiniLM) & 62.8 & 52.3 \\
 & dense (Qwen3-Emb-0.6B) & 63.8 & 53.2 \\
\midrule
\multicolumn{4}{@{}l}{\textit{Explicit memory systems}} \\
MemoryBank$^{\dagger}$
 & dense + regrouping & 56.2 & 53.9 \\
A-Mem
 & LLM + dense + links & 47.4 & 33.5 \\
Mem0$^{\dagger}$
 & dense (MiniLM) & 57.0 & 42.7 \\
LightMem
 & dense (MiniLM) & 64.0 & 51.5 \\
SimpleMem
 & LLM + hybrid + reflection & 46.4 & 29.0 \\
% Check whether "reranking" refers to a stage other than the disabled cross-encoder.
HyperMem
 & hybrid + reranking & 61.8 & 40.1 \\
\bottomrule
\end{tabular}
\caption{\textbf{RQ1: Accuracy under original system configurations.}
	Results are reported on LongMemEval-S (LME-S) and BEAM-100K (BEAM).
	$^{\dagger}$ denotes systems using the unified answer prompt; bold indicates the best result per benchmark.}
\label{tab:rq1_native}
\end{table}

We first ask if explicit memory systems outperform raw-history RAG under their original configurations. This provides an end-to-end comparison before separating the effects of explicit memory construction, retrieval, and context budget.

\paragraph{Settings.}
We compare the six explicit memory systems with the raw-history RAG baselines and full-context reference described in Section~\ref{sec:experimental_framework} on LongMemEval-S and BEAM-100K. For each explicit memory system, we preserve its original construction procedure and retrieval pipeline (Appendix~\ref{app:adaptations}). A-Mem, LightMem, SimpleMem, and HyperMem also retain their system-specific answer prompts, while MemoryBank and Mem0 use the unified answer prompt because their evaluated implementations do not define a distinct downstream answering procedure.

For raw-history RAG, we evaluate three retrieval-unit definitions, \emph{all-turns}, \emph{user-turns}, and \emph{pair-turns}, with two dense retrievers: all-MiniLM-L6-v2~\citep{all_minilm_l6_v2} and Qwen3-Embedding-0.6B~\citep{qwen3embedding}. We include all-MiniLM-L6-v2 to match the retriever used by several evaluated explicit memory systems. We also include Qwen3-Embedding-0.6B, which is used by SimpleMem, to test whether a stronger raw-history retrieval baseline changes the comparison. Each configuration retrieves the top 10 retrieval units and uses the unified answer prompt.

\vspace{0.1em}
\findingbox{Finding 1: Under original configurations, no explicit memory system outperforms the strongest raw-history RAG baseline.}
\vspace{0.1em}

The strongest raw-history RAG baseline leads on both benchmarks, although the margin is much larger on LongMemEval-S than on BEAM-100K (Table~\ref{tab:rq1_native}). In contrast, the full-context baseline performs substantially worse, suggesting that the advantage of raw history depends on selecting relevant content rather than providing more of it.

The importance of retrieval is further reflected in retriever choice. On LongMemEval-S, replacing all-MiniLM-L6-v2 with Qwen3-Embedding-0.6B improves \emph{all-turns} RAG by 14.8 percentage points and reverses its comparison with LightMem. An apparent advantage of explicit memory under a weaker raw-history baseline can therefore disappear when retrieval is strengthened, highlighting the importance of retrieval when assessing the value added by explicit memory construction.

To answer RQ1, none of the evaluated explicit memory systems outperforms the strongest raw-history RAG baseline under their original configurations. However, the comparison is sensitive to retriever choice, so these end-to-end results do not isolate the effect of explicit memory construction. A controlled retrieval comparison is therefore needed to determine whether the raw-history RAG advantage persists when retrieval conditions are aligned.
\section{RQ2: Does the Raw-History RAG Advantage Persist Under Controlled Retrieval Settings?}
\label{sec:rq2}

% Generated by _paper/scripts/build_rq_figures.py; do not edit by hand.
\begin{figure}
\centering
\includegraphics[width=0.5\textwidth]{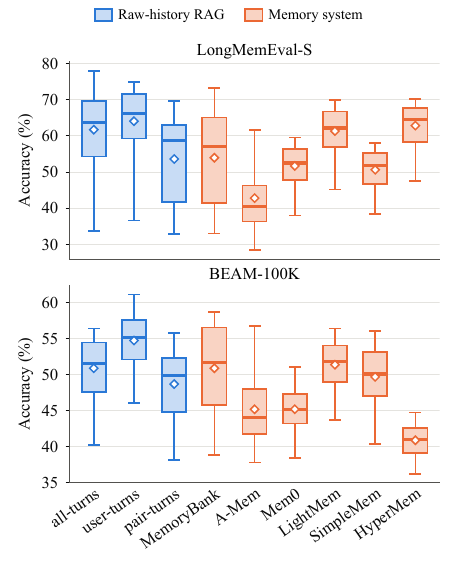}
\caption{\textbf{RQ2: Accuracy across retrievers and context budgets.}
	Each box summarizes one method over 64 configurations (8 retrievers $\times$ 8 context budgets). Boxes show the interquartile range and median, diamonds the mean, and whiskers the minimum and maximum.}
\label{fig:rq2_controlled}
\end{figure}

% Generated by _paper/scripts/build_rq_figures.py; do not edit by hand.
\begin{figure*}
\centering
\includegraphics[width=\textwidth]{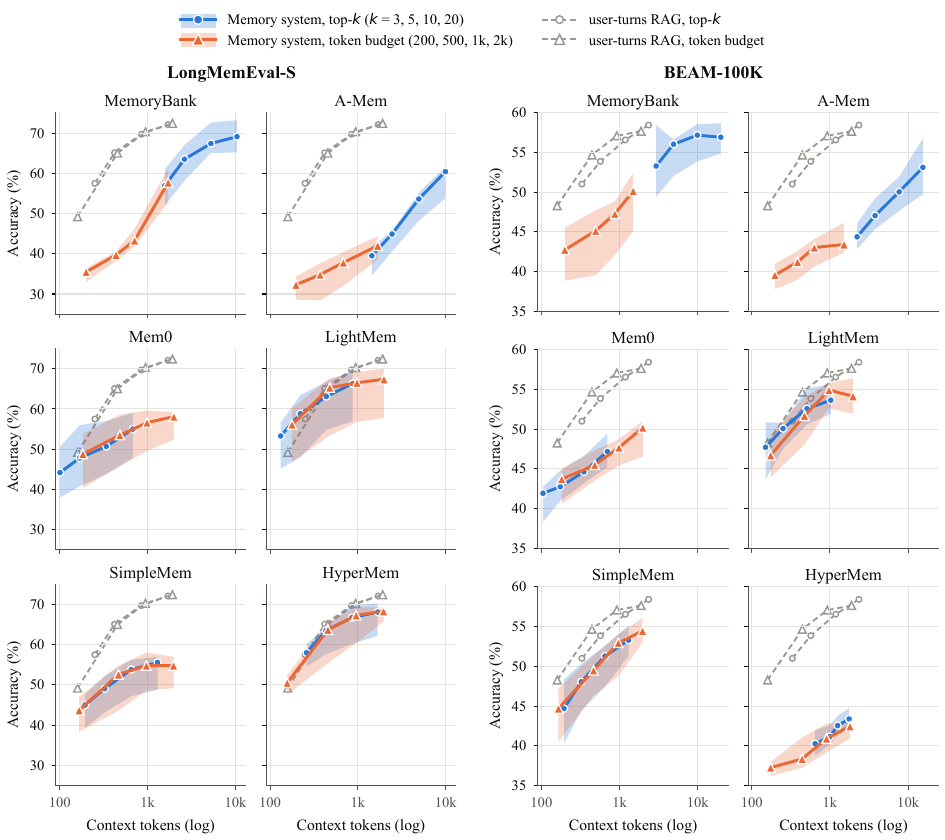}
\caption{\textbf{RQ2: Accuracy across context budgets.}
	Solid lines show the median accuracy of each explicit memory system across eight retrievers, with bands indicating the minimum and maximum. Dashed lines show the corresponding median for \emph{user-turns} RAG. The x-axis reports the number of tokens actually supplied to the answer generator.}
\label{fig:rq2_budgets}
\end{figure*}

% Generated by _paper/scripts/build_rq_tables.py; do not edit by hand.
\begin{table}[t]
\centering
\footnotesize
\setlength{\tabcolsep}{4pt}
\begin{tabular}{@{}lrrrr@{}}
\toprule
& & \multicolumn{2}{c}{Best score} & \\
\cmidrule(lr){3-4}
Question type & $n$ & RAG & Memory & $\Delta$ \\
\midrule
\multicolumn{5}{@{}l}{\textit{LongMemEval-S}} \\
Knowledge update & 72 & 88.9 & 79.2 & $-9.7$ \\
Single-session user & 64 & 98.4 & 90.6 & $-7.8$ \\
Multi-session & 121 & 74.4 & 69.4 & $-5.0$ \\
Single-session assistant & 56 & 96.4 & 92.9 & $-3.5$ \\
Single-session preference & 30 & 80.0 & 76.7 & $-3.3$ \\
Temporal reasoning & 127 & 69.3 & 68.5 & $-0.8$ \\
\cmidrule[0.2pt](lr){1-5}
Abstention & 30 & 70.0 & 76.7\rlap{\textsuperscript{a}} & {\boldmath$+6.7$} \\
\midrule
Overall & 500 & 73.2 & 69.0 & $-4.2$ \\
\midrule
\multicolumn{5}{@{}l}{\textit{BEAM-100K}} \\
Multi-session reasoning & 40 & 65.5 & 55.4 & $-10.1$ \\
Summarization & 40 & 51.6 & 45.1 & $-6.5$ \\
Instruction following & 40 & 88.8 & 84.4 & $-4.4$ \\
Temporal reasoning & 40 & 38.8 & 35.0 & $-3.8$ \\
Knowledge update & 40 & 46.2 & 43.8 & $-2.4$ \\
Preference following & 40 & 93.1 & 91.9 & $-1.2$ \\
Information extraction & 40 & 62.9 & 62.8 & $-0.1$ \\
\cmidrule[0.2pt](lr){1-5}
Event ordering & 40 & 36.0 & 37.0\rlap{\textsuperscript{b}} & {\boldmath$+1.0$} \\
Contradiction resolution & 40 & 53.8 & 56.2\rlap{\textsuperscript{c}} & {\boldmath$+2.4$} \\
Abstention & 40 & 61.2 & 76.2\rlap{\textsuperscript{d}} & {\boldmath$+15.0$} \\
\midrule
Overall & 400 & 57.7 & 54.4 & $-3.3$ \\
\bottomrule
\addlinespace[2pt]
\multicolumn{5}{@{}l}{Memory leads: \textsuperscript{a}SimpleMem, \textsuperscript{b}LightMem,} \\
\multicolumn{5}{@{}l}{\textsuperscript{c}MemoryBank and A-Mem (tie), \textsuperscript{d}HyperMem.} \\

\end{tabular}
\caption{
	\textbf{RQ2: Best accuracy by question type.}
	For each type, RAG and Memory report the best result among the three raw-history RAG baselines and six explicit memory systems, respectively, using Qwen3-Embedding-0.6B, a 1,000-token budget, and the unified answer prompt.
	$\Delta=\text{Memory}-\text{RAG}$; positive differences are bold.}

\label{tab:rq2_types}
\end{table}

% Generated by _paper/scripts/build_rq_figures.py; do not edit by hand.
\begin{figure}
\centering
\includegraphics[width=0.6\textwidth]{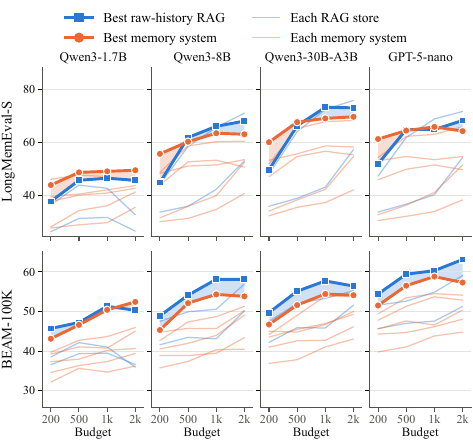}
\caption{\textbf{RQ2: Accuracy across answer generators and token budgets.}
	Thick lines highlight the raw-history RAG baseline and explicit memory system with the highest mean accuracy across these budgets.}
\label{fig:rq2_generators}
\end{figure}

Next, we test whether the raw-history RAG advantage persists under controlled retrieval settings. This allows us to assess whether the pattern observed in RQ1 is robust to changes in retriever and context budget.

\paragraph{Settings.}
We hold each system's constructed explicit memory fixed and use a unified answering setup across all explicit memory systems and raw-history RAG baselines. In the main comparison, we vary eight retrievers and eight context budgets while using Qwen3-30B-A3B-Instruct-2507 as the answer generator. The retrievers span three families: BM25~\citep{robertson2009probabilistic} and SPLADE-v3~\citep{lassance2024splade} for sparse retrieval; all-MiniLM-L6-v2, Qwen3-Embedding-0.6B, and Qwen3-Embedding-4B~\citep{qwen3embedding} for dense retrieval; and three fusion retrievers that combine BM25 with each dense retriever using reciprocal rank fusion (RRF)~\citep{cormack2009reciprocal}. As described in Section~\ref{sec:evaluation_pipeline}, we vary the retriever while preserving each system's surrounding retrieval logic (Appendix~\ref{app:rq2}).

We consider four top-$k$ budgets, $k \in \{3,5,10,20\}$, and four token budgets, $B \in \{200,500,1000,2000\}$, yielding 64 retriever and context-budget configurations per method. Under a top-$k$ budget, the answer generator receives up to the $k$ highest-ranked retrieval units. Under a token budget, retrieval units are added in rank order until adding the next unit would exceed the budget. We additionally examine performance by question type and answer generator. For the question-type analysis, we use Qwen3-Embedding-0.6B with a 1,000-token budget and compare the best raw-history RAG baseline with the best explicit memory system for each type. For the answer-generator analysis, we fix the retriever to Qwen3-Embedding-0.6B and evaluate Qwen3-1.7B, Qwen3-8B, Qwen3-30B-A3B-Instruct-2507~\citep{qwen3technicalreport}, and GPT-5-nano-2025-08-07~\citep{openai2025gpt5} across the four token budgets. Each system's constructed explicit memory and the indexed raw histories remain fixed across answer generators.

\vspace{0.1em}
\findingbox{Finding 2: The raw-history RAG advantage persists across controlled retrieval settings.}
\vspace{0.1em}

The controlled comparison preserves the main pattern from RQ1: raw-history RAG remains strongest overall across retrievers and context budgets on both benchmarks (Figure~\ref{fig:rq2_controlled}). Among the raw-history baselines, \emph{user-turns} performs best on average, while \emph{pair-turns} is weaker. This shows that the choice of retrieval-unit definition can substantially affect performance even without explicit memory construction.

The explicit memory systems are less stable across benchmarks and configurations. Systems that are competitive on LongMemEval-S can perform substantially worse on BEAM-100K, suggesting that no explicit memory construction strategy provides a consistent advantage across evaluation conditions.

Context budget also changes the comparison (Figure~\ref{fig:rq2_budgets}). Some explicit memory systems are competitive with raw-history RAG when relatively little context is supplied, especially under top-$k$ retrieval, but these gains often weaken under fixed token budgets. Top-$k$ retrieval does not equalize the amount of context supplied to the answer generator, so gains under a fixed number of retrieval units do not necessarily persist when methods receive comparable token budgets. Increasing the context budget improves some explicit memory systems, but does not consistently close the gap with raw-history RAG.

\vspace{0.1em}
\findingbox{Finding 3: Explicit memory gains are localized by question type.}
\vspace{0.1em}

The question-type analysis shows that the raw-history advantage is not driven by a single class of questions (Table~\ref{tab:rq2_types}). Raw-history RAG leads on most question types, including knowledge update and multi-session reasoning on both benchmarks. These are abilities that explicit memory construction is intended to support, indicating that constructing persistent representations does not consistently improve the use of changing or distributed information across conversations.

Abstention is the clearest recurring exception, with an explicit memory system leading on both benchmarks. Explicit memory also provides smaller gains on contradiction resolution and event ordering on BEAM-100K. These benefits are concentrated in particular abilities and arise from different systems, rather than reflecting a general advantage of explicit memory construction.

\vspace{0.1em}
\findingbox{Finding 4: Answer-generator scale changes where explicit memory helps, but not the overall pattern.}
\vspace{0.1em}

Across four answer generators, explicit memory is most competitive when model capacity or run-time context is constrained (Figure~\ref{fig:rq2_generators}; four more generators in Appendix~\ref{app:rq2}, Figure~\ref{fig:rq2_generators_app}). On LongMemEval-S, explicit memory leads across all budgets with Qwen3-1.7B and at the tightest budget for the larger generators. Within the Qwen3 family, this advantage shrinks as the answer generator becomes larger and more raw history can be supplied.

On BEAM-100K, raw-history RAG remains stronger in nearly all answer-generator and budget settings, with only an isolated reversal for Qwen3-1.7B at 2,000 tokens. The broader generator sweep shows that the raw-history RAG advantage is not specific to one answer generator or model family, while explicit memory can help more under constrained answering conditions.

To answer RQ2, the raw-history RAG advantage largely persists across common retrievers and context budgets under a unified answering setup and remains robust across answer generators. Explicit memory provides localized gains for particular question types and under constrained run-time context or answer-generator capacity, but these gains do not persist consistently across benchmarks or settings.
\section{RQ3: Is Explicit Memory Limited by Construction or Retrieval?}
\label{sec:rq3}

Even under the controlled retrieval settings in RQ2, explicit memory systems can remain behind raw-history RAG. However, this comparison does not reveal where the limitation arises. A system may underperform because explicit memory construction does not preserve the information needed for answering, or because the information is preserved but the retrieval pipeline fails to return it. We therefore use \emph{oracle retrieval} to distinguish these two sources of limitation.

\paragraph{Settings.}
Oracle retrieval bypasses the practical retriever and uses gold annotations to identify the information that should be supplied to the answer generator. We conduct this analysis on BEAM-100K because its annotations identify the conversation turns supporting each question. For raw-history RAG, we directly supply the retrieval units corresponding to these supporting turns. For explicit memory systems, we trace the supporting turns through explicit memory construction and supply the resulting memory units (Appendix~\ref{app:rq3}). All methods use the same unified answering setup. Oracle accuracy therefore measures how well each representation can support answering when practical retrieval errors are removed.

\begin{figure}
\centering
\includegraphics[width=0.6\textwidth]{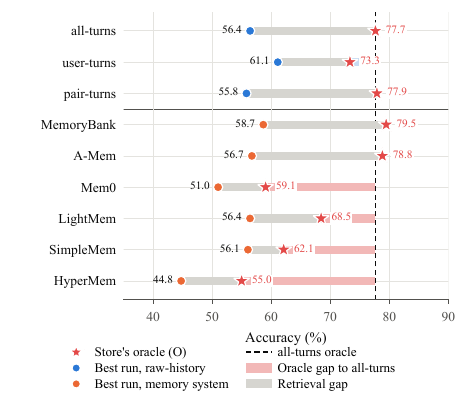}
\caption{\textbf{RQ3: Oracle versus retrieved accuracy on BEAM-100K.}
	$\star$ denotes oracle accuracy and $\bullet$ denotes accuracy under the best retrieval setting from RQ2. The dashed line marks the \emph{all-turns} oracle. Pink bars show the oracle gap to \emph{all-turns}, and grey bars show the gap between oracle accuracy and the best retrieval setting.}
\label{fig:rq3_beam_oracle}
\end{figure}

Figure~\ref{fig:rq3_beam_oracle} compares each method's \emph{oracle accuracy} with its \emph{best retrieved accuracy}, defined as its highest answer accuracy across the 64 retriever and context-budget configurations in RQ2. We examine two gaps. For explicit memory systems, the gap between oracle accuracy and the \emph{all-turns} oracle indicates how much answer-supporting capability remains relative to raw history after explicit memory construction. The gap between oracle and best retrieved accuracy indicates how much of that capability is not realized by practical retrieval.

% Generated by _paper/scripts/build_rq_figures.py; do not edit by hand.

\vspace{0.1em}
\findingbox{Finding 5: Most explicit memory systems remain below raw history even under oracle retrieval.}
\vspace{0.1em}

Bypassing practical retrieval does not eliminate the gap between explicit memory and raw history (Figure~\ref{fig:rq3_beam_oracle}; per-type results in Appendix~\ref{app:rq3}, Table~\ref{tab:rq3_oracle_types}). MemoryBank and A-Mem approach the \emph{all-turns} oracle, whereas the other four explicit memory systems remain substantially below it. For these systems, improving retrieval alone cannot close the gap, pointing instead to limitations in the constructed explicit memory.

\vspace{0.1em}
\findingbox{Finding 6: High oracle potential is often not realized by practical retrieval.}
\vspace{0.1em}

High oracle accuracy does not necessarily translate into high retrieved accuracy. MemoryBank and A-Mem approach the \emph{all-turns} oracle, yet their best retrieved accuracy remains approximately 21--22 percentage points lower. A similarly large oracle-to-retrieved gap for \emph{all-turns} RAG shows that this retrieval bottleneck is not unique to explicit memory systems.

Oracle potential and practical retrieval performance are therefore distinct. \emph{user-turns}, for example, has lower oracle accuracy than \emph{all-turns} but achieves higher best retrieved accuracy, showing that lower oracle potential can still yield better practical performance when the available information is easier to retrieve. Conversely, a small oracle-to-retrieved gap does not necessarily indicate effective retrieval when oracle accuracy itself is low.

To answer RQ3, explicit memory systems exhibit two distinct bottlenecks. Some are limited by the information available after explicit memory construction, while others preserve substantial answer-supporting potential that practical retrieval fails to realize. Improving conversational memory therefore requires both preserving useful information during construction and making that information reliably retrievable.
\section{RQ4: Does Explicit Memory Justify Its Additional Computational Cost?}
\label{sec:rq4}

% Generated by _paper/scripts/build_rq_tables.py; do not edit by hand.
\begin{table*}[t]
	\centering
	\footnotesize
	\setlength{\tabcolsep}{4pt}
	\begin{tabular*}{\textwidth}{@{\extracolsep{\fill}}l@{\hskip 14pt}rrr@{\hskip 14pt}rrrrrr@{\hskip 14pt}r@{}}
		\toprule
		& \multicolumn{3}{c}{Construction} & \multicolumn{2}{c}{Retrieval} & \multicolumn{2}{c}{Answer} & \multicolumn{2}{c}{Overall} & \\
		\cmidrule(lr){2-4}\cmidrule(lr){5-6}\cmidrule(lr){7-8}\cmidrule(lr){9-10}
		System & Calls & Tokens (k) & Min & Tok/q (k) & s/q & Tok/q (k) & s/q & Tok/q (k) & s/q & Acc.\ (\%) \\
		\midrule
		\multicolumn{11}{@{}l}{\textit{LongMemEval-S}} \\
		Full context
		& 0 & 0 & 0 & 0 & 0 & 95.7 & 8.7 & 95.7 & 8.7 & 22.2 \\
		\emph{all-turns}
		& 0 & 0 & 0 & 0 & 0.014 & 1.8 & 1.1 & 1.8 & 1.1 & \textbf{78.0} \\
		\emph{user-turns}
		& 0 & 0 & 0 & 0 & 0.010 & 1.1 & 1.0 & 1.1 & 1.0 & 72.0 \\
		\emph{pair-turns}
		& 0 & 0 & 0 & 0 & 0.010 & 4.7 & 1.4 & 4.7 & 1.4 & 63.8 \\
		MemoryBank
		& 47.8 & 134.1 & 8.3 & 0 & 0.007 & 3.2 & 3.3 & 3.2 & 3.3 & 56.2 \\
		A-Mem
		& 987.7 & 2,718.4 & 215.1 & 0.1 & 0.323 & 15.7 & 2.6 & 15.8 & 2.9 & 47.4 \\
		Mem0
		& 417.6 & 697.6 & 41.7 & 0 & 0.015 & 7.0 & 1.6 & 7.0 & 1.6 & 57.0 \\
		LightMem
		& 108.6 & 113.9 & 6.2 & 0 & 0.007 & 0.7 & 0.6 & 0.7 & 0.6 & 64.0 \\
		SimpleMem
		& 33.6 & 297.7 & 24.4 & 10.3 & 6.956 & 4.3 & 2.1 & 14.5 & 9.1 & 46.4 \\
		HyperMem
		& 363.5 & 1,802.7 & 87.9 & 0 & 1.388 & 9.2 & 2.5 & 9.2 & 3.9 & 61.8 \\
		\midrule
		\multicolumn{11}{@{}l}{\textit{BEAM-100K}} \\
		Full context
		& 0 & 0 & 0 & 0 & 0 & 96.3 & 11.8 & 96.3 & 11.8 & 24.4 \\
		\emph{all-turns}
		& 0 & 0 & 0 & 0 & 0.008 & 3.9 & 1.3 & 3.9 & 1.3 & \textbf{56.0} \\
		\emph{user-turns}
		& 0 & 0 & 0 & 0 & 0.008 & 1.7 & 1.0 & 1.7 & 1.0 & 55.9 \\
		\emph{pair-turns}
		& 0 & 0 & 0 & 0 & 0.011 & 10.2 & 1.8 & 10.2 & 1.8 & 53.2 \\
		MemoryBank
		& 4.5 & 136.1 & 0.9 & 0 & 0.006 & 6.2 & 1.8 & 6.2 & 1.8 & 53.9 \\
		A-Mem
		& 573.2 & 2,160.4 & 116.2 & 0.2 & 0.854 & 26.7 & 3.4 & 27.0 & 4.3 & 33.5 \\
		Mem0
		& 271.6 & 648.1 & 25.6 & 0 & 0.013 & 0.6 & 0.9 & 0.6 & 0.9 & 42.7 \\
		LightMem
		& 101.9 & 100.0 & 3.6 & 0 & 0.006 & 1.0 & 0.9 & 1.0 & 0.9 & 51.5 \\
		SimpleMem
		& 19.4 & 259.8 & 11.1 & 10.4 & 7.206 & 5.1 & 1.7 & 15.4 & 8.9 & 29.0 \\
		HyperMem
		& 19.6 & 280.7 & 7.3 & 0 & 0.693 & 7.0 & 2.4 & 7.0 & 3.1 & 40.1 \\
		\bottomrule
	\end{tabular*}
	\caption{\textbf{RQ4: Cost by stage.}
		Construction reports per-conversation LLM calls, tokens, and wall time;
		Retrieval and Answer report per-question cost.
		Overall is the sum of Retrieval and Answer.
		Tokens include LLM input and output; retrieval tokens are zero when no
		LLM is used at query time. Bold indicates the best score per dataset.}
	\label{tab:rq4_cost}
\end{table*}

So far, we have focused on answer accuracy, but explicit memory systems can also introduce substantial computation during explicit memory construction and at query time. Construction is an upfront cost, whereas query-time cost recurs for every question, so the value of an explicit memory system depends on both its computational cost and resulting answer accuracy. RQ4 therefore asks whether the additional computation required to construct and use explicit memory is justified by its downstream benefits.

\paragraph{Settings.}
We measure computational cost at three stages: explicit memory construction, retrieval, and answer generation. Each explicit memory system uses its original configuration, as in RQ1. For raw-history RAG, we use Qwen3-Embedding-0.6B with top-10 retrieval units and the unified answer prompt.

Construction cost is measured per conversation, averaged across each dataset, using LLM calls, LLM tokens, and wall time. Retrieval and answer-generation costs are measured per question using LLM tokens and wall time. Token counts include both input and output tokens, and all timing measurements use an H100 GPU (Appendix~\ref{app:cost}). Table~\ref{tab:rq4_cost} reports these costs alongside answer accuracy. Query-time cost combines retrieval and answer generation and excludes construction.

\vspace{0.1em}
\findingbox{Finding 7: Explicit memory construction can incur substantial upfront cost.}
\vspace{0.1em}

Explicit memory construction can require substantial computation before any questions are answered. On LongMemEval-S, even LightMem, the fastest system to construct, requires 6.2 minutes per conversation, while A-Mem requires approximately 35 times longer. Raw-history RAG avoids this LLM-based construction stage entirely. The large variation across systems shows that construction cost depends strongly on the explicit memory construction procedure. Evaluating only query-time cost therefore misses part of the computational cost of explicit memory systems.

\vspace{0.1em}
\findingbox{Finding 8: Additional computation does not consistently translate into higher answer accuracy.}
\vspace{0.1em}

Across systems, higher computational cost does not consistently correspond to higher answer accuracy. Several explicit memory systems incur substantial construction cost and higher query-time cost than raw-history RAG while still achieving lower accuracy. Additional computation during explicit memory construction and retrieval therefore does not consistently translate into better downstream performance.

Lower query-time cost is possible, but it does not necessarily offset the accuracy trade-off. For example, LightMem uses fewer query-time LLM tokens than \emph{all-turns} RAG on LongMemEval-S but is 14.0 percentage points less accurate. Moreover, query-time savings do not require explicit memory construction: on BEAM-100K, \emph{user-turns} achieves nearly the same accuracy as \emph{all-turns} while using substantially fewer query-time LLM tokens and requiring no LLM-based memory construction. Thus, lower query-time cost is neither unique to explicit memory nor sufficient by itself to establish an efficiency advantage.

To answer RQ4, the evaluated explicit memory systems do not consistently convert their additional computation into higher answer accuracy. Some reduce query-time cost, but these savings must be considered together with their accuracy trade-offs and upfront construction cost. Whether construction cost is worthwhile also depends on how many future queries it can be amortized over, but within the evaluated settings, explicit memory does not show a consistent cost--accuracy advantage over raw-history RAG.
\section{Discussion and Conclusion}

In this study, we asked whether explicit memory construction provides enough benefit to justify its additional cost over raw-history RAG. Across six state-of-the-art explicit memory systems, we find no consistent overall advantage in answer accuracy over strong raw-history RAG baselines, with gains limited to particular question types and tight run-time context budgets. Oracle retrieval reveals two distinct limitations: some systems fail to preserve enough answer-supporting information during construction, while others preserve useful information that practical retrieval fails to access. These limited gains are further offset by substantial construction and, in some cases, query-time computation that does not consistently improve answer accuracy.

These results suggest that long-term conversational memory should not be viewed only and primarily as a problem of constructing increasingly elaborate explicit memory. Conversation history already provides a persistent source of information, and the value of explicit memory construction depends on whether it preserves answer-supporting information and makes that information easier to retrieve within the run-time context budget. Construction and retrieval must therefore be considered jointly: better retrieval cannot recover information unavailable after construction, while richer explicit memory provides little benefit if its useful information cannot be retrieved.

Looking forward, raw-history RAG should serve as a strong reference point rather than merely a simple baseline. A promising direction is to retain raw history while using explicit memory construction selectively when abstraction, consolidation, or structure provides measurable downstream value. Future work should therefore focus not only on how to construct richer explicit memory, but on when such construction is useful, what information should be transformed, and whether its gains in answer accuracy or information access justify the additional computational cost.

\bibliography{../references} % arXiv uses the bundled .bbl

% Appendix, assembled after the references (TACL appendix policy, March 2024).
% Category 1, replication details: limit 5 pages; prose allowed.
% Category 2, complementary results: limit 3 pages; tables and figures only,
% each referenced from the main text.
% Nothing here may carry a finding the main text relies on.
% Start directly after the references (TACL only requires appendices to follow them).
\appendix
% Top-align float-only pages here and keep a fixed gap between their floats
% (LaTeX centres them and stretches the gaps by default).
\makeatletter
\setlength{\@fptop}{0pt}
\setlength{\@dblfptop}{0pt}
\setlength{\@fpsep}{16pt}
\setlength{\@dblfpsep}{16pt}
\makeatother

% Sections mirror the main text: Section 3, then RQ1--RQ4. Extended related work is free text outside
% Category 1, so TACL does not allow it here. Do not restate the main text.

% Mirrors Section 3. Category 1: replication details.

\section{Experimental Framework}
\label{app:framework}

\subsection{Systems and Baselines}
\label{app:systems}

Table~\ref{tab:app_adaptations} summarizes the construction, retrieval, and answering components of each explicit memory system.

\paragraph{Source code.}
We use each system's public implementation and integrate all systems through external adapters. For Mem0, we use version v0.1.118.

\paragraph{Construction adaptations.}
For MemoryBank, we summarize each session separately, raise the summary output cap from 400 to 1{,}024 tokens, and disable forgetting. For A-Mem, we raise the construction output cap from 1{,}000 to 4{,}000 tokens. For Mem0, we replace text-embedding-3-small with all-MiniLM-L6-v2 and assign facts their session dates. For SimpleMem, we cap extraction outputs at 16{,}000 tokens and retain the public implementation's setting with JSON mode disabled.

\paragraph{Raw-history unit formatting.}
The \emph{all-turns} and \emph{pair-turns} baselines retain the \texttt{user:} and \texttt{assistant:} prefixes, whereas \emph{user-turns} omits them. Each pair joins a user turn and the following assistant turn with a newline.

\paragraph{Full-context baseline.}
The \emph{full-context} baseline bypasses retrieval and supplies raw conversation history directly to the answer generator. We cap the history at 100K tokens and, when necessary, retain the most recent content that fits.

\begin{table*}[t]
	\centering
	\footnotesize
	\setlength{\tabcolsep}{3.5pt}
	\renewcommand{\arraystretch}{1.18}
	\newcommand{\nt}[1]{\textsuperscript{\,#1}}
	\begin{tabularx}{\textwidth}{@{}>{\raggedright\arraybackslash}p{2.1cm}*{6}{>{\raggedright\arraybackslash}X}@{}}
		\toprule
		& \textbf{MemoryBank} & \textbf{A-Mem} & \textbf{Mem0} & \textbf{LightMem} & \textbf{SimpleMem} & \textbf{HyperMem} \\
		\midrule
		\multicolumn{7}{@{}l}{\textit{Construction}} \\
		Input & Session & Turn & User--assistant pair & User turns, topic-segmented & 20-turn window, 5-turn overlap & Session \\
		Retrieval unit & Session summary; turn pair & Note with keywords, tags, links & Fact & Fact sentence & Atomic fact with metadata & Topic, episode, fact \\
		Key step & Summarize session & Link to $k{=}5$ neighbors; memory evolution\nt{a} & Extract, then add / update / delete\nt{b} & LLMLingua-2 compression, rate 0.8; offline update, $\theta{=}0.8$\nt{c} & Extract per window & Extract episodes and facts; group by topic \\
		\midrule
		\multicolumn{7}{@{}l}{\textit{Retrieval}} \\
		Query-time LLM & -- & Keyword query & -- & -- & Query plan; ${\le}2$ reflections & -- \\
		Retriever / scorer & MiniLM\nt{d} & MiniLM & MiniLM & MiniLM & Qwen3-Emb-0.6B, keyword, SQL & RRF: BM25 + Qwen3-Emb-4B \\
		Retrieval depth & 6 & 10 & 200 & 10 & 25 / 5 / 5\nt{e} & 100 $\rightarrow$ 10 / 10 / 30\nt{f} \\
		Post-processing & Regroup by source\nt{g} & One-hop link expansion & -- & -- & Merge legs & Qwen3-Reranker-4B \\
		\midrule
		\multicolumn{7}{@{}l}{\textit{Answering}} \\
		Answer prompt & Unified & System-specific & Unified & System-specific & System-specific & System-specific (chain of thought) \\
		\bottomrule
	\end{tabularx}
\caption{\textbf{Configurations of explicit memory systems.}
		Construction components are used in all experiments; retrieval and answering components are those of the original configurations used in RQ1 (Section~\ref{sec:rq1}) and RQ4 (Section~\ref{sec:rq4}); -- denotes none. Qwen3-Emb abbreviates Qwen3-Embedding.
		$^{a}$An LLM may update the new note's links and tags and its neighbors' context.
		$^{b}$An LLM compares each new fact with similar stored facts.
		$^{c}$Similarity threshold for rewriting an earlier fact.
		$^{d}$all-MiniLM-L6-v2.
		$^{e}$Semantic / keyword / structured legs.
		$^{f}$Candidates, then topics / episodes / facts.
		$^{g}$Hits from the same session are concatenated.
	~\vspace{-10pt}}
	\label{tab:app_adaptations}
	
\end{table*}

\subsection{Datasets and Evaluation}
\label{app:judging}

\paragraph{Judge settings.}
We use GPT-4o mini through the OpenAI Batch API with temperature~0, seed~42, and a 1{,}024-token output cap. For LongMemEval-S, we retain the official question-type-specific prompts and separate abstention prompt, which request yes/no verdicts. For BEAM-100K's rubric-based questions, the judge assigns each criterion 0, 0.5, or 1, and the question score is the mean across its criteria.

\paragraph{Event alignment.}
For BEAM-100K event ordering, we change only the event-matching procedure. The released implementation evaluates each response line against each reference event separately and accepts the first match. We instead send all numbered response lines and all shuffled, labeled reference events in one judge call and request a one-to-one mapping.

\paragraph{BEAM-100K judge prompt.}
We restore the question field shown in the BEAM paper's rubric prompt but omitted from the released judging implementation. We also clarify that explicitly stating the absence of relevant information is responsive, while awarding credit only when the rubric criterion itself requires that absence. The excerpt below shows the prompt's opening, input fields, and complete responsiveness block.

\begin{promptcard}{BEAM-100K rubric judge prompt}
	{\normalfont\scriptsize\itshape Relevant excerpt. Original BEAM text is
		unshaded; blue shading marks our added abstention clarification; gray ellipses
		indicate unchanged omitted text.}\par\smallskip
	\promptrole{User}
	You are an expert evaluator tasked with judging whether the LLM\textquotesingle{}s response demonstrates compliance with the specified RUBRIC CRITERION.
	\par\smallskip
	\#\# EVALUATION INPUTS\\
	- QUESTION (what the user asked): \promptslot{<question>}\\
	- RUBRIC CRITERION (what to check): \promptslot{<rubric\_item>}\\
	- RESPONSE TO EVALUATE: \promptslot{<llm\_response>}
	\promptskip
	\#\# RESPONSIVENESS REQUIREMENT (anchored to the QUESTION)\\
	A compliant response must be **on-topic with respect to the QUESTION** and attempt to answer it.\\
	- If the response does not address the QUESTION, score **0.0** and stop.
	\begin{promptchanged}
		- An explicit statement that the conversation contains no such information IS a way of addressing the QUESTION, not a failure to address it. Treat such a response as responsive and score it against the RUBRIC CRITERION as normal; it satisfies the criterion only if the criterion is itself about that absence.
	\end{promptchanged}
	- For negative constraints, both must hold: (a) the response is responsive to the QUESTION, and (b) the prohibited element is absent.
	\promptskip
\end{promptcard}

\subsection{Models and Answering Setup}
\label{app:prompts}

\paragraph{Unified answer prompt.}
The unified answer prompt follows LightMem's LongMemEval prompt, with the question date and question on separate lines. Each retrieved unit is preceded by its session date. We omit the date line when no date is available and render an empty retrieval result as \emph{(no relevant memories found)}.

\begin{promptcard}{Unified answer prompt}
	\promptrole{System}
	You are a helpful assistant.
	\tcblower
	\promptrole{User}
	Question time: \promptslot{\{question\_date\}}\\
	Question: \promptslot{\{question\}}\par\smallskip
	Please answer the question based on the following memories:\\
	{[}\promptslot{\{timestamp\}}{]} \promptslot{\{retrieved unit text\}}\\
	{[}\promptslot{\{timestamp\}}{]} \promptslot{\{retrieved unit text\}}\\
	\textcolor{gray!70}{\ldots}
\end{promptcard}

\paragraph{Decoding settings.}
Where supported, LLM calls use temperature~0. Answer generation uses seed~42 and a 2{,}000-token output cap, except HyperMem, which uses 4{,}096 tokens. Qwen3-1.7B, Qwen3-8B, and Qwen3-32B run in non-thinking mode. GPT-5-nano uses minimal reasoning effort and does not expose a temperature setting. GPT-5-nano and GPT-4o mini are accessed through the OpenAI API.
% Mirrors Section 4. Category 1: replication details.
\section{RQ1: Original Configurations}
\label{app:adaptations}

RQ1 and RQ4 use each system's original retrieval and answering components (Table~\ref{tab:app_adaptations}). Appendix~\ref{app:rq1_details} details these configurations, while Appendix~\ref{app:rq1_construction_model} examines the sensitivity of the results to the choice of memory construction model.

\subsection{Implementation Details}
\label{app:rq1_details}

\paragraph{Retrieval.}
We remove duplicate notes returned by A-Mem's link expansion. Mem0's depth of 200 follows its LongMemEval evaluation. SimpleMem uses the fixed depths of its public implementation, which does not implement the adaptive depth described in its paper.

\paragraph{System-specific answer prompts.}
The templates of A-Mem, SimpleMem, and HyperMem omit the question date. HyperMem's template requests chain-of-thought reasoning.

\subsection{Sensitivity to the Memory Construction Model}
\label{app:rq1_construction_model}

\begin{table}[t]
\centering
\footnotesize
\setlength{\tabcolsep}{2pt}
\begin{tabular}{@{}lrrr@{}}
\toprule
& \multicolumn{2}{c}{Construction model} & \\
\cmidrule(lr){2-3}
Memory system & Qwen3-30B-A3B & GPT-4o mini & $\Delta$ \\
\midrule
MemoryBank & 53.9 & 54.3 & $+0.4$ \\
A-Mem & 33.5 & 28.8 & $-4.7$ \\
Mem0 & 42.7 & 54.2 & $+11.5$ \\
LightMem & 51.5 & 51.1 & $-0.4$ \\
SimpleMem & 29.0 & 31.8 & $+2.8$ \\
HyperMem & 40.1 & 34.7 & $-5.4$ \\
\midrule
\multicolumn{4}{@{}l}{\textit{Raw-history RAG (reference)}} \\
\emph{all-turns} (MiniLM) & \multicolumn{2}{c}{\textbf{56.2}} & -- \\
\bottomrule
\end{tabular}
\caption{
	\textbf{RQ1: Accuracy of explicit memory systems across memory construction models.}
	Results are reported on BEAM-100K.
	$\Delta=\text{GPT-4o mini}-\text{Qwen3-30B-A3B}$; bold indicates the best result under both construction models.
	The bottom row is the strongest raw-history RAG baseline from Table~\ref{tab:rq1_native}~\vspace{-3pt}}
\label{tab:rq1_construction_model}
\end{table}

Section~\ref{sec:rq1} uses Qwen3-30B-A3B-Instruct-2507 for explicit memory construction. To test whether Finding~1 depends on this choice, we reconstruct the explicit memory of each system on BEAM-100K with GPT-4o mini while keeping its original retrieval and answering setup. The two runs therefore differ only in the LLM used for explicit memory construction, which we refer to as the construction model.

For most explicit memory systems, changing the construction model shifts accuracy by only a few percentage points (Table~\ref{tab:rq1_construction_model}). Mem0 is the exception and gains substantially with GPT-4o mini. These shifts do not change the RQ1 comparison: under either construction model, no explicit memory system outperforms the strongest raw-history RAG baseline, so Finding~1 holds.

% Mirrors Section 5. Category 1 details, then Category 2 exhibits.
\section{RQ2: Controlled Retrieval Settings}
\label{app:rq2}

\paragraph{Retriever implementation.}
We implement the eight retrievers used in RQ2 independently of the explicit memory systems. all-MiniLM-L6-v2 and Qwen3-Embedding-0.6B are also used for the raw-history RAG baselines in RQ1. BM25 uses \texttt{rank\_bm25} with default parameters and lowercased word tokens. Qwen3-Embedding queries use the instruction \emph{``Given a query, retrieve relevant passages that answer the query''}, while documents receive no instruction. Fusion combines the complete BM25 and dense rankings using
$\mathrm{score}(d)=\sum_{\ell}1/(60+\mathrm{rank}_\ell(d)+1)$.
Each conversation is indexed separately, without retriever fine-tuning.

\paragraph{Retriever replacement.}
As in the main experiments, we replace the retriever while preserving each system's surrounding retrieval logic. In MemoryBank, the replacement retriever ranks turn pairs and session summaries before units from the same session are regrouped. In A-Mem, it ranks notes using the generated keyword query before one-hop link expansion. In Mem0 and LightMem, it replaces the native dense-retrieval stage. In SimpleMem, it replaces semantic retrieval in each sub-query and reflection round, while query planning, keyword and structured retrieval, merging, and reflection remain unchanged. In HyperMem, it ranks candidates at the topic, episode, and fact layers, while retaining the cross-encoder reranker.

\paragraph{Token budgets.}
Token budgets count rendered retrieval units using the answer generator's tokenizer, including timestamp prefixes and one token per newline separator, but excluding the question and answer-prompt instructions. Units are added in rank order and are never skipped to fit a later unit. If the first unit alone exceeds the budget, we truncate it and retain its prefix.

\begin{figure}
\centering
\includegraphics[width=0.6\textwidth]{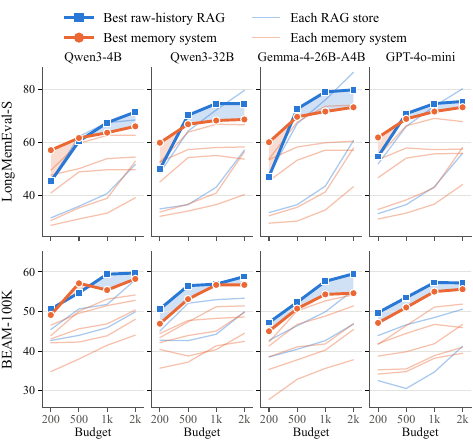}
\caption{
	\textbf{RQ2: Accuracy across four additional answer generators.}
	Qwen3-4B, Qwen3-32B, Gemma-4-26B-A4B~\citep{gemmateam2026gemma4}, and GPT-4o mini, with Qwen3-Embedding-0.6B as the retriever. Layout as in Figure~\ref{fig:rq2_generators}.
}
\label{fig:rq2_generators_app}
\end{figure}

\paragraph{Additional answer generators.}
Figure~\ref{fig:rq2_generators_app} extends the answer-generator analysis with four additional models. We keep Qwen3-Embedding-0.6B as the retriever and use the same four token budgets as in the main analysis. Each system's constructed explicit memory and the indexed raw histories remain fixed across answer generators.

% Generated by _paper/scripts/build_rq_figures.py; do not edit by hand.

% TODO: Construction backbone -- build memory with Gemma-4 and answer with Qwen3
% (and the reverse), retriever fixed. Needs new runs.
% (If added, restore subsections: Retrievers and Context Budgets / Construction Models.)
% \subsection{Construction Models}
% \label{app:cross_backbone}
% Mirrors Section 6. Category 1 details, then Category 2 exhibits.
\section{RQ3: Oracle Retrieval}
\label{app:rq3}

Table~\ref{tab:rq3_oracle_types} reports oracle accuracy by question type on BEAM-100K using the oracle contexts defined below.

\begin{table*}[t]
\centering
\footnotesize
\setlength{\tabcolsep}{1pt}
\begin{tabular*}{\textwidth}{@{\extracolsep{\fill}}l*{9}{r}@{}}
\toprule
& \multicolumn{3}{c}{Raw-history RAG} & \multicolumn{6}{c}{Explicit memory systems} \\
\cmidrule(lr){2-4}\cmidrule(l){5-10}
\textbf{Question type} & \emph{all-turns} & \emph{user-turns} & \emph{pair-turns} & MemoryBank & A-Mem & Mem0 & LightMem & SimpleMem & HyperMem \\
\midrule
Contradiction resolution & \textbf{61.9} & 60.0 & 51.6 & 48.4 & 56.6 & 54.1 & 60.9 & 48.8 & 47.2 \\
Event ordering & 71.3 & 74.8 & 68.4 & 70.0 & \textbf{76.6} & 33.9 & 57.6 & 28.1 & 30.6 \\
Information extraction & \textbf{88.9} & 79.9 & 88.5 & 88.5 & 88.2 & 71.2 & 75.6 & 71.5 & 47.9 \\
Instruction following & 77.5 & 81.2 & 86.2 & \textbf{97.5} & 90.0 & 78.1 & 73.1 & 76.9 & 81.2 \\
Knowledge update & \textbf{72.5} & 63.8 & 68.8 & 71.2 & \textbf{72.5} & 32.5 & 53.8 & 41.2 & 37.5 \\
Multi-session reasoning & 75.8 & 76.3 & 79.1 & \textbf{82.3} & 82.1 & 44.8 & 75.5 & 63.0 & 52.7 \\
Preference following & 91.9 & \textbf{95.6} & \textbf{95.6} & 95.0 & 95.0 & 85.0 & 92.5 & 91.2 & 85.6 \\
Summarization & 75.5 & 43.0 & \textbf{79.5} & 78.0 & 69.8 & 62.2 & 43.1 & 61.7 & 51.0 \\
Temporal reasoning & 61.2 & 58.8 & 61.2 & \textbf{63.8} & 57.5 & 28.8 & 52.5 & 38.8 & 16.2 \\
Abstention & 100.0 & 100.0 & 100.0 & 100.0 & 100.0 & 100.0 & 100.0 & 100.0 & 100.0 \\
\midrule
Overall & 77.7 & 73.3 & 77.9 & \textbf{79.5} & 78.8 & 59.1 & 68.5 & 62.1 & 55.0 \\
\bottomrule
\end{tabular*}
\caption{
	\textbf{RQ3: Oracle accuracy by question type on BEAM-100K.}
	Each question type contains 40 questions; Overall reports accuracy across all 400 questions, including those with empty oracle contexts. Bold indicates the best result per row, except for abstention, where all methods tie.
}
\label{tab:rq3_oracle_types}
\end{table*}

\paragraph{Oracle selection.}
Each retrieval unit records its \emph{provenance}, namely the source turn or span from which it was constructed. We include a retrieval unit in the oracle context when its provenance overlaps at least one annotated supporting turn. Provenance is an individual turn for \emph{all-turns}, \emph{user-turns}, A-Mem, and LightMem; a turn pair for \emph{pair-turns} and Mem0; a 20-turn window for SimpleMem; and a full session for HyperMem. MemoryBank traces each turn-pair memory to its two constituent turns and each session summary to its full session. We apply no top-$k$ or token budget and use an empty context when no retrieval unit matches. Because a window or session may contain information beyond the annotated evidence, provenance overlap identifies potentially relevant units but does not guarantee that they preserve the information needed to answer the question.
% Mirrors Section 7. Category 1: replication details.
\section{RQ4: Cost Measurement}
\label{app:cost}

\paragraph{Environment.}
Each LLM-serving job uses a single 80\,GB H100 GPU with vLLM v0.22.1. Answer generation runs with concurrency 1.

\paragraph{Construction.}
Construction timing covers the LLM calls used to construct explicit memory for one conversation and excludes index embedding. Construction token counts include both input and output tokens from these calls.

\paragraph{Retrieval.}
Retrieval timing covers the full call to $P_s$, including query encoding, search, and any query-time LLM calls, such as A-Mem's keyword generation and SimpleMem's query planning and reflection. It excludes model loading, explicit memory loading, and index construction. Retrieval token counts include only input and output tokens from query-time LLM calls, not embedding operations.

\paragraph{Answer generation.}
Answer-generation timing begins when a request is sent and ends when its response is received. Token counts include both input and output tokens. For HyperMem, output tokens include the chain-of-thought reasoning requested by its system-specific answer prompt.

\end{document}